\documentclass{article}

\PassOptionsToPackage{numbers, compress}{natbib}

\usepackage[main, final]{neurips_2026}

\usepackage[utf8]{inputenc}
\usepackage[T1]{fontenc}
\usepackage{microtype}

\usepackage{amsmath,amssymb,amsfonts,amsthm}
\usepackage{mathtools}
\usepackage{nicefrac}
\usepackage{bm}              
\usepackage{pgfplots}
\pgfplotsset{compat=1.18}

\usepackage{tabularray}
\usepackage{placeins}

\usepackage{algorithm}
\usepackage{algpseudocodex}

\usepackage[table]{xcolor}
\definecolor{sectiongray}{gray}{0.88}

\usepackage{booktabs}         
\usepackage{multirow}
\usepackage{array}

\usepackage{graphicx}
\usepackage{subcaption}       
\usepackage{tikz}
\usetikzlibrary{arrows.meta, positioning, calc, shapes.geometric, fit}

\usepackage{xcolor}
\definecolor{cuphblue}{HTML}{4477AA}      
\definecolor{cmpred}{HTML}{EE6677}        
\definecolor{crispercolor}{HTML}{AA3377}  
\definecolor{gudhigray}{HTML}{BBBBBB}     
\definecolor{accentgreen}{HTML}{228833}   

\usepackage{hyperref}
\hypersetup{
  colorlinks=true,
  linkcolor=cuphblue,
  citecolor=cuphblue,
  urlcolor=cuphblue,
  bookmarksnumbered=true,
}
\usepackage{url}
\usepackage{cleveref}         

\usepackage{enumitem}         
\usepackage{wrapfig}          
\usepackage{etoolbox}         

\title{TRiST: Topology-Reinforced Swin Transformer for Medical Image Analysis}

\author{
\textbf{Pengfei Gu}$^{1}$ \qquad
\textbf{Huimin Li}$^{2}$ \qquad
\textbf{Guangyu Meng}$^{3}$ \qquad
\textbf{Hao Zheng}$^{4}$
\\[0.4em]
\textbf{Haoteng Tang}$^{1}$ \qquad
\textbf{Bin Fu}$^{1}$ \qquad
\textbf{Danny Z. Chen}$^{3}$
\\[0.7em]
$^{1}$Department of Computer Science, The University of Texas Rio Grande Valley\\
$^{2}$School of Mathematical and Statistical Sciences,\\
The University of Texas Rio Grande Valley\\
$^{3}$Department of Computer Science and Engineering, University of Notre Dame\\
$^{4}$School of Computing and Informatics, University of Louisiana at Lafayette\\
{\tt pengfei.gu01@utrgv.edu}
}
\begin{document}
\maketitle

\begin{abstract}
Topological patterns in medical images, such as connected components
and loops, across multiple spatial scales carry critical structural
information, from microaneurysm-scale lesions to organ-level boundaries.
Despite the success of hierarchical Vision Transformers  (e.g., Swin
Transformer) in medical image analysis, existing methods lack an explicit
architectural scheme to represent and incorporate multi-scale topological
structures through the Transformer hierarchy.
In this paper, we propose \emph{Topology-Reinforced Swin Transformer (TRiST)},
a new Swin Transformer model family for medical image analysis, which reinforces
multi-scale hierarchical topology into the hierarchical Transformer
architecture. Instead of treating topology as an auxiliary descriptor or
post hoc fusion cues, TRiST takes topology as a native part
of representation construction, refinement, and token interaction.
First, because topology does not compose hierarchically through pooling or
interpolation, we develop a Hierarchical Multi-Scale Topology
Representation 
algorithm that computes 2D persistent
homology ($H_0$ and $H_1$ on image patches) at each 
native scale of Swin Transformer, constructing four topological representations aligned 1-to-1 with the Swin
token grid.
Second, motivated by the persistence lifetime of topological
features, we develop a Band-Aware $H_0/H_1$ Residual
Refinement module that separately processes short-,
medium-, and long-persistence information using
dedicated gated residual MLPs.
Third, we introduce a Persistence-Band Progressive
Attention Bias that converts stage-specific topological
descriptors into learned key-saliency scores and
injects them into shifted-window attention logits.
Based on these strategies, we instantiate two task-specific models:
Topo-SwinV2-B for classification and Topo-Swin-UNet for segmentation.
Both models retain the original Swin-based backbone structure while equipping it
with native multi-scale topological features.
Experiments on five medical image datasets 
show consistent
improvements over Swin-based baselines and competitive performance relative to existing topology-augmented methods.
\end{abstract}

\section{Introduction}
\label{sec:intro}

Medical image analysis benefits largely from models that jointly capture fine-grained
local appearance and large-scale structural organization~\cite{lawson2019persistent,adame2025topo,an2025hierarchical,zhang2024ihcsurv,liang2026cell}.
In many medical imaging applications,
spatial organization and structural
patterns provide complementary
information to photometric and
semantic representations.
In diabetic retinopathy grading, microaneurysms manifest as small isolated
connected components whose number and spatial distribution across retinal
quadrants determine the disease severity~\cite{mishra2021vtg,li2025topology}.
In dermoscopy, malignant lesions exhibit irregular boundary loops and
asymmetric connected structures that distinguish them from benign nevi~\cite{gu2026integrating, meng2026topocl,meng2026topoagent,meng2026vtrl}.
In cardiac MRI, the myocardium forms a closed annular boundary whose
topological integrity is clinically meaningful~\cite{gu2025self}.
In mammography, suspicious microcalcification clusters form spatially
distributed point sets whose connectivity and persistence distinguish
malignant from benign cases~\cite{peng2024phg}.
In these examples, the relevant signals span multiple spatial scales:
fine-scale topology captures local lesion structures, intermediate-scale
topology captures vascular connectivity and tissue boundaries, and
coarse-scale topology captures global anatomical enclosures and
organ-level distributions.
Such topological properties, e.g., connected components and loops
captured by $H_0$ and $H_1$ persistent homology (PH)~\cite{edelsbrunner2002topological} 
on 2D image
patches and their birth and death across filtration thresholds, are not captured well by pixel intensity alone.
An explicit multi-scale representation of topological
structures can largely boost the capability of deep learning (DL) models. 

Despite the strong inductive bias that topology can provide for medical
imaging, integrating PH
into DL models remains an under-explored topic.
Standard DL methods for medical image classification and
segmentation (e.g., ResNet~\cite{resnet},
SwinV2-B~\cite{liu2022swin},
U-Net~\cite{ronneberger2015u},
TransUNet~\cite{chen2021transunet},
Swin-UNet~\cite{cao2022swin},
Attention U-Net~\cite{oktay2018attention},
and UNETR~\cite{hatamizadeh2022unetr}) operate purely on visual features with no representation of
topological persistence.
Topology-guided loss approaches~\cite{hu2019topology,hu2020topology,
clough2020topological,shit2021cldice,he2023toposeg} incorporate
topological constraints into training objectives to encourage
topologically correct predictions, but do not inject topology into
intermediate feature representations; the model must still infer
topological structures from visual features alone.
Topology-as-features approaches learn from persistence diagrams (PDs)~\cite{adams2017persistence} directly
or fuse persistence descriptors with visual features.
PersLay~\cite{perslay} and PLLay~\cite{pllay} introduce learnable
vectorization layers for PDs; PersFormer~\cite{reinauer2021persformer}
embeds persistence features into Transformer architectures;
PHG-Net~\cite{peng2024phg} and its multi-scale extension~\cite{gu2026integrating} fuse multi-scale topological descriptors with CNN features
for medical image classification.
TopoTxR~\cite{wang2025topotxr} extracts 1D (loop) and 2D (bubble)
topological structures from breast DCE-MRI using PH
and incorporates them into a 3D CNN with a topology-guided spatial
attention mechanism, directing the model's focus toward biologically
relevant voxels near fibroglandular tissue structures for pathological
complete response prediction.
While TopoTxR shows that topological structures can meaningfully
guide spatial attention, it operates at a single spatial scale,
applies topology as an external attention mask rather than an
additive bias inside the attention logit, and is designed for a
3D CNN rather than a hierarchical Vision Transformer.
More broadly, known topology-as-features methods \cite{perslay,pllay,
reinauer2021persformer,wang2024multiset,kim2024xpert,gu2025topoimages,meng2026topocl,meng2026topoagent, 
gu2026integrating,hofer2017deep,peng2024phg} typically
fuse topology at a single scale or as a global descriptor, losing
the spatial localization and scale correspondence essential for
dense prediction tasks.
More importantly, these methods do not compute topology at
multiple scales natively and directly from raw images, and do not place topology \emph{inside}
the attention computation of a hierarchical Vision 
Transformer to directly modulate which tokens communicate with one another at each spatial scale.

In this work, we aim to show that hierarchical Vision Transformers, and in particular Swin
Transformer~\cite{liu2022swin}, are naturally compatible with multi-scale
topology computation and integration.
By applying shifted-window self-attention at four progressively coarser
spatial resolutions, the Swin hierarchy
defines four distinct spatial scales, each with a well defined token
stride and spatial support.
This hierarchy provides a principled scaffold for multi-scale topology:
computing topology at the four native resolutions of the Swin hierarchy
ensures that each topological representation is spatially aligned with the corresponding Swin
feature map by construction, without requiring any resizing, pooling, or
interpolation.
However, exploiting this compatibility requires overcoming two non-trivial obstacles: (1) How to compute topological representation at each scale correctly and (2) how to inject
topology into the attention mechanism in order to directly modulate
token communication.

We propose \textbf{Topology-Reinforced Swin Transformer (TRiST)},
a new Swin Transformer model family, which reinforces multi-scale hierarchical topology in the hierarchical Transformer
architecture.
Our design is guided by the following observations.
\emph{(1) Topology does not compose hierarchically.}
A simplest approach to compute multi-scale topology is to compute a single
topology representation at the finest resolution and downsample it to coarser stages
via pooling or interpolation.
However, this is fundamentally flawed: the PD of a large region
captures qualitatively different topological patterns than the average of
many small-region PDs, because global structures such as
vessel loops and organ enclosures may become complete only at an appropriate
spatial scale.
Thus, downsampling a fine-scale topological representation can destroy
scale-specific semantic content that makes topology comprehensive.
We develop a {\em Hierarchical Multi-Scale Topology
Representation} algorithm that computes PH
independently at each stage-specific native scale, producing four
topology representations that are aligned 1-to-1 with the Swin token grid at each
stage.
\emph{(2) The persistence axis of PD has semantic meaning.}
Note that features near the diagonal of a PD have short lifetime
and correspond to ephemeral events such as noise and fine texture, while
features far from the diagonal have long lifetime and encode stable
structures such as organ boundaries, vessel walls, and lesion clusters.
Treating all persistence elements uniformly with a single MLP conflates
noise-level features with structurally significant ones, and can lead the
model to inadvertently suppress diagnostically relevant long-lifetime
features.
We design a \emph{Band-Aware $H_0/H_1$ Residual Refinement}
module that independently refines each of six persistence lifetime
sub-bands (i.e., $H_0$ and $H_1$ each split into short, medium, and long
persistence) using dedicated gated residual MLPs, respecting both the
semantic structure of the persistence axis and the geometric distinction
between connectivity ($H_0$) and enclosure ($H_1$).
\emph{(3) Topologically salient tokens should attract stronger
attention.}
Tokens corresponding to microaneurysms, vessel junctions, or boundary
segments have high topological salience, and their local persistence
descriptors encode structurally significant events relative to
surrounding background tokens.
Standard shifted-window self-attention
learns content-dependent token interactions
but does not explicitly incorporate
persistence-derived structural salience.
We propose a \emph{Persistence-Band Progressive Attention
Bias} that adds an outer-sum topology bias
to the attention logits.
After row-wise softmax, the query-dependent component
of the outer-sum bias cancels, leaving an effective
topology-conditioned key-saliency prior.
Consequently, tokens assigned higher topology scores
can contribute more strongly to the updated
representations of neighboring query tokens.

We instantiate our TRiST framework using two specific architectures: {Topo-SwinV2-B} for medical image classification built on
Swin Transformer~\cite{liu2022swin}, and {Topo-Swin-UNet} for medical
image segmentation built on Swin-UNet~\cite{cao2022swin}.
Both models share the same topology pipeline and require no structural changes to
the original backbones.
On benchmark datasets that include dermoscopy classification (ISIC 2018), mammography
classification (CBIS-DDSM), retinal grading (RetinaMNIST), multi-organ CT
segmentation (Synapse), and cardiac MRI segmentation (ACDC), both architectures consistently improve over their strong Swin-based baselines
and achieve competitive performance relative to topology-augmented
methods, with only negligible additional parameter costs.

\begin{figure}[t]
    \centering
    \includegraphics[width=1.0\linewidth]{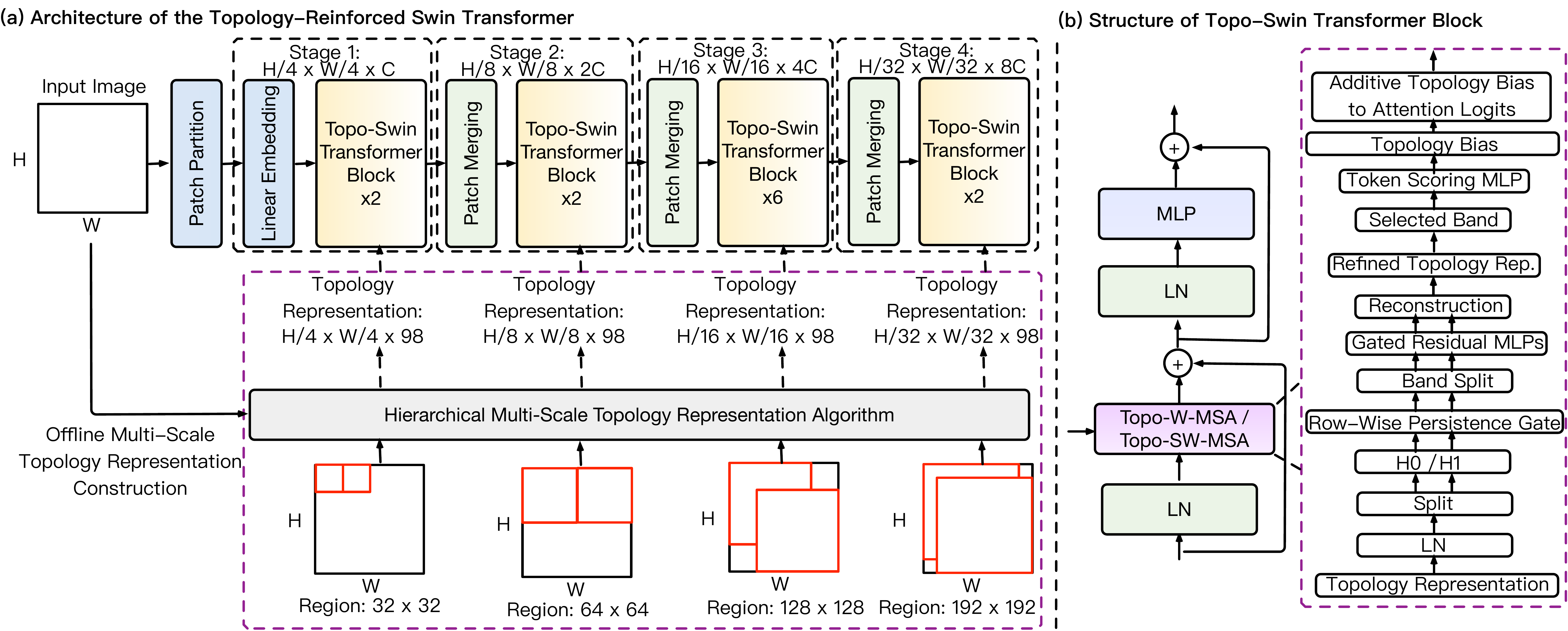}
    \caption{(a) The overall architecture of our Topology-Reinforced Swin Transformer (TRiST). In parallel to the Swin backbone, we pre-compute hierarchically aligned multi-scale topological representations at four native scales. (b) The structure of a TRiST block. Each stage-specific topological representation is normalized and refined by the proposed Band-Aware $H_0/H_1$ Residual Refinement module, and then decomposed into persistence bands to construct an additive topology bias for window attention. The resulting topology bias is injected into the attention logits together with the relative position bias. 
Stages 1, 2, and 3--4 use short, medium, and long bands, respectively.
    }
    \label{fig:architecture}
\end{figure}

\section{Method}
\label{sec:method}

TRiST has three main
components (see Fig.~\ref{fig:architecture}): (1) a hierarchical multi-scale topology representation
algorithm that pre-computes topology at four native Swin resolutions;
(2) a Band-Aware $H_0/H_1$ Residual Refinement module that adapts 
topology representations to downstream tasks; (3) a
Persistence-Band Progressive Attention Bias module that injects
stage-specific topology 
into shifted-window self-attention.
Sect.~\ref{sec:application} presents the classification/segmentation instantiation models.

\subsection{Hierarchical Multi-Scale Topology Representation}
\label{sec:topo_scales}

Let $\mathbf{I}\in\mathbb{R}^{H\times W\times 3}$ be an input 2D image. For a Swin
backbone with patch stride $p=4$, the spatial size of the stage-$s$
feature map is
$$
H_s \times W_s
=
\frac{H}{p\,2^{s-1}} \times \frac{W}{p\,2^{s-1}},
$$
and the corresponding token stride in the original image is
$
S_s = p\,2^{s-1},
s\in\{1,2,3,4\}.
$
Each token at stage $s$ therefore corresponds to a spatial footprint of
$S_s \times S_s$ pixels in the original image, growing from $4\times4$
pixels at Stage~1 to $32\times32$ pixels at Stage~4.

A simple way to obtain topology at multiple scales is to compute one
topology representation at the finest resolution and downsample it in the coarser stages via average pooling or bilinear interpolation.
But, topology does not compose hierarchically under
pooling or interpolation.
Pooling a persistence image (PI)~\cite{adams2017persistence} averages the birth--death pair densities
of small local regions, but it does not produce the PD of a
larger region.
The PD of a $64\times64$ region captures qualitatively
different topological events than the average of sixteen $16\times16$
PDs: the former may detect a single closed vessel loop
spanning the large region, while the latter can only aggregate point-wise
statistics of much smaller structures.
Thus, resizing a topology representation can conflate scale-specific topological semantics, destroying the correspondence between a descriptor and the spatial scale that it aims to represent.

To address this issue, we compute PH separately at each
scale on image regions whose size follows the same progression as the
Swin stage hierarchy.
For stage $s$, a topology descriptor (e.g., PI) is extracted from an image region
of size
$
    R_s \;=\; S_s\,k_s,
$
where $k_s$ is an overlap multiplier controlling the spatial context assigned
to each token-aligned descriptor.
We use
$
    (k_1,k_2,k_3,k_4) \;=\; (8,8,8,6),
$
yielding region sizes $32\times32$, $64\times64$, $128\times128$, and
$192\times192$ for Stages~1--4, respectively.
Stages~1--3 share the same multiplier $k_s=8$, maintaining a consistent
context ratio across scales.
At Stage~4, using $k_4=8$ would produce a $256\times256$ region, exceeding
the $224\times224$ image boundary; reducing $k_4$ to 6 yields a $192\times192$
region, whose area is approximately 73.5\%
of the standardized $224\times224$ image area.
Boundary-centered crops may still extend beyond
the image domain and are handled by reflection
padding.
To center each topology region on the corresponding Swin token, the image is
reflect-padded by
$
    \mathrm{pad}_s
    \;=\;
    \left\lfloor \frac{R_s}{2} \right\rfloor
    -
    \left\lfloor \frac{S_s}{2} \right\rfloor
$
pixels on each side before region extraction.
The region extracted at token position $(i,j)$ is then centered at the same
image location as the corresponding stage-$s$ Swin token.
\begin{algorithm}[t]
\caption{Hierarchical Multi-Scale Topology Representation}
\label{alg:topo_pipeline}
\small
\begin{algorithmic}[1]

\Require Input image $\mathbf{I}\in\mathbb{R}^{H\times W\times 3}$,
         patch stride $p=4$,
         overlap multipliers $(k_1,k_2,k_3,k_4)=(8,8,8,6)$,
         vectorization operator $\mathcal{V}$

\Ensure $\{\mathbf{P}_s\}_{s=1}^{4}$ with
        $\mathbf{P}_s\in\mathbb{R}^{H_s\times W_s\times 98}$
        aligned with the stage-$s$ token grid

\For{$s=1$ \textbf{to} $4$}
    \State $S_s \gets p\,2^{s-1}$, \quad $R_s \gets S_s\,k_s$,
           \quad $H_s \gets H/S_s$, \quad $W_s \gets W/S_s$
    \State $\mathrm{pad}_s \gets \lfloor R_s/2 \rfloor - \lfloor S_s/2 \rfloor$;
           \quad $\mathbf{I}^{\mathrm{pad}} \gets \mathrm{ReflectPad}(\mathbf{I},\,\mathrm{pad}_s)$
    \State Initialize $\mathbf{P}_s\in\mathbb{R}^{H_s\times W_s\times 98}$
    \For{each position $(i,j)$ on the $H_s\times W_s$ grid}
        \State Extract $R_s\times R_s$ crop centered at token $(i,j)$
               from $\mathbf{I}^{\mathrm{pad}}$
        \State Convert to grayscale (if color crop); max-normalize to $[0,1]$
        \State Compute $\mathrm{PD}^{H_0}_{s,i,j}$ and
               $\mathrm{PD}^{H_1}_{s,i,j}$ via sublevel-set
               persistent homology; 
        \State $\mathbf{v}^{H_0}_{s,i,j}\gets
               \mathrm{vec}\bigl(\mathcal{V}(\mathrm{PD}^{H_0}_{s,i,j})\bigr)
               \in\mathbb{R}^{49}$
        \State $\mathbf{v}^{H_1}_{s,i,j}\gets
               \mathrm{vec}\bigl(\mathcal{V}(\mathrm{PD}^{H_1}_{s,i,j})\bigr)
               \in\mathbb{R}^{49}$
        \State $\mathbf{P}_s[i,j,:]\gets
               [\,\mathbf{v}^{H_0}_{s,i,j};\,\mathbf{v}^{H_1}_{s,i,j}\,]$
               \Comment{ [\,;\,\,] is concatenation}
    \EndFor
\EndFor
\State \Return $\{\mathbf{P}_1,\mathbf{P}_2,\mathbf{P}_3,\mathbf{P}_4\}$

\end{algorithmic}
\end{algorithm}

For each stage $s$ and token position $(i,j)$,
we extract an $R_s\times R_s$ image crop.
For RGB inputs, grayscale intensity is computed as
$
I_{\mathrm{gray}}
=
0.299R+0.587G+0.114B.
$
Each crop is then normalized by its maximum
intensity:
\begin{equation}
\widehat{I}(x)
=
\begin{cases}
I_{\mathrm{gray}}(x)/
\max_y I_{\mathrm{gray}}(y),
& \max_y I_{\mathrm{gray}}(y)>0,\\
0, & \text{otherwise}.
\end{cases}
\end{equation}
We employ a sublevel-set filtration
$
\mathcal{F}_t =
\{x : \widehat{I}(x) \leq t\},
$
in which lower-intensity pixels enter
before higher-intensity pixels on the
corresponding cubical complexes, obtaining two PDs:
$\mathrm{PD}^{H_0}_{s,i,j}$ for connected components ($H_0$) and
$\mathrm{PD}^{H_1}_{s,i,j}$ for loops ($H_1$).
The essential $H_0$ class (with infinite death) is removed before
vectorization.
Let $\mathcal{V}(\cdot)$ denote a diagram vectorization operator.
In our implementation, each vectorized descriptor is reshaped to a $7\times7$ grid, yielding a 49-dimensional vector per homology group.
The stage-$s$ topology descriptor at position $(i,j)$ is
$$
   \mathbf{P}_s[i,j,:]
\;=\;
\bigl[\,
    \mathbf{v}^{H_0}_{s,i,j};\,
    \mathbf{v}^{H_1}_{s,i,j}
\,\bigr]
\;\in\; \mathbb{R}^{98},
$$
where vectors $\mathbf{v}^{H_0}_{s,i,j},\mathbf{v}^{H_1}_{s,i,j}\in\mathbb{R}^{49}$.
Collecting all positions yields
$\mathbf{P}_s\in\mathbb{R}^{H_s\times W_s\times 98}$, spatially
aligned with the Swin visual feature map at stage $s$.
The 
construction procedure is given in
Algorithm~\ref{alg:topo_pipeline}; stage-wise region sizes and topology 
dimensions are summarized in Appendix Table~\ref{tab:topo_scales}.
Note that the pipeline for multi-scale topology representation construction is descriptor-agnostic
at the PD level: the same region extraction,
sublevel-set homology computation, and four-scale alignment mechanism
are applied regardless of the vectorization chosen.
The vectorization operator $\mathcal{V}$ can be PI,
persistence landscapes (PL)~\cite{bubenik2015statistical},
Betti curves (BC)~\cite{chung2022persistence}, or
persistence statistics (PS)~\cite{ali2023survey}
without modifying the hierarchical alignment mechanism.

Native multi-scale topology computation preserves the semantics of the Swin
hierarchy: early stages capture local topological structures, while later
stages capture larger and more stable anatomical patterns. By
computing topology at each stage-specific region size, the resulting topology
representations remain both scale-appropriate and spatially aligned with the Swin
hierarchy (see Table~\ref{tab:descriptor}).

\subsection{Band-Aware $H_0/H_1$ Residual Refinement}
\label{sec:band_refine}

The raw topology representation $\mathbf{P}_s$ is computed directly from pixel intensities and is thus task-agnostic.
A useful refinement module must satisfy two requirements simultaneously:
it must preserve the semantic structure already encoded by PH, and it must adapt that structure to the downstream task without destabilizing the pretrained Swin backbone.
Hence, we refine each stage-specific topology representation using a structured residual module that respects two principles:
(i) $H_0$ and $H_1$ encode qualitatively different geometries and should not be mixed;
(ii) persistence lifetime carries meaningful semantics and should not be processed by a single MLP over all 98 dimensions.

Before refinement, each stage-specific topology representation is normalized along its descriptor dimension:
$
    \bar{\mathbf{P}}_s
    \;=\;
    \mathrm{LN}(\mathbf{P}_s),
$
where $\mathrm{LN}$ denotes a LayerNorm over the 98 channels.
We split $\bar{\mathbf{P}}_s$ into its two homology groups:
$$
    \bar{\mathbf{P}}_s
    \;=\;
    \bigl[
        \bar{\mathbf{P}}^{(H_0)}_s
        \;\|\;
        \bar{\mathbf{P}}^{(H_1)}_s
    \bigr],
\bar{\mathbf{P}}^{(H_0)}_s,\bar{\mathbf{P}}^{(H_1)}_s\in\mathbb{R}^{H_s\times W_s\times 49}.
$$
Each 49-dimensional vector is reshaped into a $7\times7$ grid:
$
    \mathbf{X}^{(g)}_s
    \;\in\;
    \mathbb{R}^{H_s\times W_s\times 7\times 7},
    g\in\{H_0,H_1\}.
$
For PIs, the rows
of the $7\times7$ grid correspond to
different persistence-lifetime ranges.
We partition these rows into short
(rows 0--1), medium (rows 2--4), and
long (rows 5--6) persistence bands.
For BC, PL, and PS, we retain the same
grid organization and index partitions
for architectural consistency. However,
these descriptors do not generally have
the same intrinsic birth--persistence
geometry as PIs, so their groups are
fixed index partitions rather than
literal persistence-lifetime intervals.



This design is motivated by two facts: persistence lifetime has semantic
meaning, and $H_0$ and $H_1$ encode different geometries. Thus, we
avoid processing all 98 dimensions with a single shared MLP; instead, we
refine each homology group and persistence band separately.


\noindent
\textbf{Row-wise Persistence Gate.}
Before splitting into three bands, we apply a learnable row-wise gate to the $7\times7$ topology grid for each homology group:
$$
    \widetilde{\mathbf{X}}^{(g)}_s
    \;=\;
    \mathbf{X}^{(g)}_s
    \odot
    \Bigl(
        \mathbf{1} + \sigma(\boldsymbol{\gamma}^{(g)}_s)
    \Bigr),
$$
where $\boldsymbol{\gamma}^{(g)}_s\in\mathbb{R}^{7}$ is a learnable vector broadcast over the column axis and all spatial positions, $\odot$ is Hadamard multiplication, and
$\sigma(\cdot)$ is the sigmoid function.
This gate provides a stage-specific, globally shared reweighting of persistence lifetime before the fixed short/medium/long split is applied.

\noindent
\textbf{Band-specific Residual Refinement.}
For each stage $s$, homology group $g\in\{H_0,H_1\}$, and band
$b\in\{\mathrm{short},\mathrm{medium},\mathrm{long}\}$, we flatten the corresponding slice of $\widetilde{\mathbf{X}}^{(g)}_s$ into a vector
$\mathbf{z}^{(g,b)}_s$ and refine it with a dedicated two-layer MLP:
$$
    \widehat{\mathbf{z}}^{(g,b)}_s
    \;=\;
    \mathbf{z}^{(g,b)}_s
    +
    \sigma(\alpha^{(g,b)}_s)\,
    \mathrm{MLP}^{(g,b)}_s\!\left(\mathbf{z}^{(g,b)}_s\right),
$$
where $\alpha^{(g,b)}_s\in\mathbb{R}$ is a learnable scalar gate.
Each $\mathrm{MLP}^{(g,b)}_s$ has hidden dimension 32 and is parameterized independently.
The short and long bands have input dimension 14 per homology group, while the medium band has input dimension 21 per homology group.
To preserve the pretrained backbone during early training, the output layer of every band-specific MLP is zero-initialized, and the residual gate is initialized with
$
    \alpha^{(g,b)}_s = -6,
$
so that $\sigma(\alpha^{(g,b)}_s)\approx 0.0025$.
The zero-initialized output layers and
small residual gates ensure that the
band-specific MLP updates are initially
negligible, stabilizing topology refinement
during fine-tuning.
However, LayerNorm and row-wise gating
still transform the raw topology descriptors.
Thus, the near-identity initialization applies
to the residual MLP updates rather than to
the entire topology refinement pipeline.


\noindent
\textbf{Reconstruction.}
After refining all the six sub-bands, we reconstruct the refined stage-$s$ descriptor by concatenating them in their original order:
\begin{equation*}
    \widetilde{\mathbf{P}}_s[i,j,:]
    \;=\;
    \Bigl[
        \widehat{\mathbf{z}}^{(H_0,\mathrm{short})}_s,\;
        \widehat{\mathbf{z}}^{(H_0,\mathrm{medium})}_s,\;
        \widehat{\mathbf{z}}^{(H_0,\mathrm{long})}_s,\;
        \widehat{\mathbf{z}}^{(H_1,\mathrm{short})}_s,\;
        \widehat{\mathbf{z}}^{(H_1,\mathrm{medium})}_s,\;
        \widehat{\mathbf{z}}^{(H_1,\mathrm{long})}_s
    \Bigr]
    \in \mathbb{R}^{98}.
\end{equation*}
A separate refinement module is used at each stage, yielding four independent instances that adapt the topology representations to the visual semantics of their corresponding Swin stages.

The band-specific residual updates provide a
controlled mechanism for adapting the
topology descriptors during fine-tuning.
Although the residual MLPs are initialized
with zero output, LayerNorm and row-wise
gating still transform the original descriptors.
The refinement therefore starts with
negligible residual updates rather than
an identity transformation of the raw
topology representation.


\subsection{Persistence-Band Progressive Attention Bias}
\label{sec:pbpab}
Standard feature-fusion approaches incorporate
topological descriptors into visual representations
through concatenation, addition, or other fusion
operations. When applied before self-attention,
such approaches may also influence attention
weights indirectly through learned query and
key projections. However, in medical images,
topologically salient tokens may provide useful
structural information to other tokens within
the same attention window. To explicitly
emphasize these tokens during attention,
we introduce a learned topology-conditioned
additive bias into the pre-softmax attention
logits, allowing selected key tokens to
contribute more strongly to neighboring
query representations.

\noindent
\textbf{Stage-wise Band Selection.}
From the refined stage-specific topology
representation $\widetilde{\mathbf{P}}_s$,
we select a persistence band according to
the following progressive routing strategy:
\begin{equation*}
    \mathbf{B}_s
    \;=\;
    \begin{cases}
        \mathrm{Short}(\widetilde{\mathbf{P}}_1), & s=1,\\
        \mathrm{Medium}(\widetilde{\mathbf{P}}_2), & s=2,\\
        \mathrm{Long}(\widetilde{\mathbf{P}}_s), & s\in\{3,4\},
    \end{cases}
\end{equation*}
where $\mathbf{B}_1\in\mathbb{R}^{H_1\times W_1\times 28}$,
$\mathbf{B}_2\in\mathbb{R}^{H_2\times W_2\times 42}$, and
$\mathbf{B}_3,\mathbf{B}_4\in\mathbb{R}^{H_s\times W_s\times 28}$.
We adopt a progressive band-routing strategy
motivated by the hierarchical organization
of Swin Transformer. Short-, medium-, and
long-persistence bands are emphasized in
Stages~1, 2, and 3--4, respectively.
Importantly, persistence lifetime and spatial
support are distinct quantities. A small crop
may contain highly persistent features,
whereas a large crop may contain short-lived
features. Thus, the proposed routing is an
architectural inductive bias rather than
an assumed correspondence between
persistence lifetime and spatial scale.
Its effectiveness relative to uniform-band
alternatives is evaluated in Table~\ref{tab:descriptor}.

\noindent
\textbf{Token Scoring.}
For every stage, a lightweight scoring network maps the selected band vector at each token to a per-head topology score:
$$
    \mathbf{s}_{s}(i,j)
    \;=\;
    \mathrm{MLP}^{\mathrm{score}}_s\!\left(\mathbf{B}_s(i,j)\right)
    \in
    \mathbb{R}^{h_s},
$$
where $h_s$ is the number of attention heads at stage $s$.
The scoring MLP has the form
$\mathrm{Linear}(d_s,64)\rightarrow\mathrm{GELU}\rightarrow\mathrm{Linear}(64,h_s)$,
where $d_s\in\{28,42\}$ is the band dimension.
Its output layer is zero-initialized so that the topology bias is initially negligible.

\noindent
\textbf{Window-aligned Topology Bias.}
The score map is padded, cyclically shifted, and partitioned into windows using exactly the same operations as the visual feature map in shifted-window attention.
This alignment is essential: the topology bias must correspond to the same token pairs that appear in the stage-specific attention matrix.
Within each window, the topology bias is constructed by an outer-sum form:
$$
    \mathbf{B}^{\mathrm{topo}}_s(u,v)
    \;=\;
    \sigma(\beta_s)\,
    \bigl(
        \mathbf{s}_{s}(u) + \mathbf{s}_{s}(v)
    \bigr),
$$
where $u$ and $v$ index two tokens in the same window and $\beta_s$ is a learnable scalar gate.
In implementation, $\beta_s$ is initialized as $-5$, so that $\sigma(\beta_s)\approx 0.0067$ and the bias begins near zero.

\noindent
\textbf{Attention-logit Injection.}
Let $\mathbf{A}_s$ be the pre-softmax attention logits for a stage-$s$ window.
For Swin-V1 blocks, the topology-reinforced attention logits are
$$
    \mathbf{A}_s
    \;=\;
    \frac{\mathbf{Q}\mathbf{K}^{\top}}{\sqrt{d}}
    +
    \mathbf{B}^{\mathrm{rel}}_s
    +
    \mathbf{B}^{\mathrm{topo}}_s
    +
    \mathbf{M}_s,
   $$
where $\mathbf{M}_s$ is the shifted-window attention mask.
For Swin-V2 blocks, the dot-product term is replaced by cosine attention
with a learned logit scale, while the topology bias is injected at the
same location:
$$
    \mathbf{A}_s
    \;=\;
    \tau_s\,
    \frac{\mathbf{Q}}{\lVert \mathbf{Q} \rVert}
    \frac{\mathbf{K}^{\top}}{\lVert \mathbf{K} \rVert}
    +
    \mathbf{B}^{\mathrm{rel}}_s
    +
    \mathbf{B}^{\mathrm{topo}}_s
    +
    \mathbf{M}_s,
$$
where $\tau_s$ is a learned logit scale.

For a fixed query token $u$, the term
$\sigma(\beta_s)\mathbf{s}_s(u)$
is constant across all keys and therefore
cancels under row-wise softmax.
Consequently, the effective topology-dependent
contribution is
$\sigma(\beta_s)\mathbf{s}_s(v)$,
which acts as a learned key-saliency prior.
For fixed visual attention logits, a key assigned
a higher topology score receives a larger relative
attention weight. Its associated value representation
can therefore contribute more strongly to the
updated representations of query tokens within
the same window.
Thus, the proposed mechanism explicitly modulates
attention using input-derived persistent-homology
descriptors, but it does not implement a fully
pairwise topological query--key interaction.


\subsection{Applications to Medical Image Classification and Segmentation}
\label{sec:application}

We instantiate our TRiST framework in two settings: Topo-SwinV2-B for classification
and Topo-Swin-UNet for segmentation. In both cases, stage-aligned
topology is pre-computed, refined, and converted into stage-specific
attention biases. For classification, we additionally use a lightweight
topology bypass branch to Swin Transformer V2 Base~\cite{liu2022swin} for preserving global topology cues. For
segmentation, the same topology pipeline is used within a Swin-UNet~\cite{cao2022swin} encoder-decoder architecture. Full architectural details of these models are provided in Appendix Sect.~\ref{appd:application}.

\section{Experiments}

\textbf{Datasets.}
We evaluate our TRiST approach on three medical image classification datasets, ISIC 2018~\cite{codella2019skin},
CBIS-DDSM~\cite{lee2017curated}, and RetinaMNIST~\cite{medmnistv2}, and two medical image segmentation datasets, Synapse~\cite{landman2015miccai} and
ACDC~\cite{bernard2018deep}. Dataset details are provided in Appendix Sect.~\ref{apped:dataset}.

\subsection{Main Results}

\textbf{Classification.}
Table~\ref{tab:cls} gives the classification results on the ISIC 2018,
CBIS-DDSM, and RetinaMNIST datasets. We compare standard DL classification baselines, including ResNet152, SENet154, and SwinV2-B, two topology-based methods PHG-Net and its multi-scale extension (PHG-Net + multi-scale), and four variants of our proposed Topo-SwinV2-B trained with multi-scale topology representations produced by different descriptors: PI, BC, PS, and PL.
From Table~\ref{tab:cls}, one can see: (1) Standard baselines perform competitively on
ISIC 2018 but are weaker on the more challenging CBIS-DDSM and
RetinaMNIST datasets, suggesting that visual features alone are
insufficient for capturing subtle structural cues. (2) Topology-based
methods improve over the standard baselines, confirming the usefulness
of topology. (3) Among all methods, the proposed Topo-SwinV2-B variants
achieve the best or near-best performance across the three datasets. On
CBIS-DDSM, Topo-SwinV2-B w/ PI achieves the best accuracy (75.62),
while PHG-Net + multi-scale attains a slightly higher AUC (82.34 vs.\
82.01). On ISIC 2018 and RetinaMNIST, our proposed variants achieve the
best results in all reported metrics. Some GradCAM visualizations are given in Appendix Sect.~\ref{apped:qual}.

\begin{table}[t]
    \centering
    \caption{Experimental results of different methods on the medical image classification datasets.} 
    \label{tab:cls}
    \scalebox{0.83}{
    \begin{tabular}{lcccccc}
        \toprule
        & \multicolumn{2}{c}{ISIC2018~\cite{codella2019skin}} & \multicolumn{2}{c}{CBIS-DDSM~\cite{lee2017curated}} & \multicolumn{2}{c}{RetinaMNIST~\cite{medmnistv2}} \\
        \cmidrule(lr){2-3}\cmidrule(lr){4-5}\cmidrule(lr){6-7}
        Method  & Acc $\uparrow$& AUC $\uparrow$& Acc $\uparrow$& AUC $\uparrow$& Acc $\uparrow$& AUC $\uparrow$ \\
        \midrule
        ResNet152~\cite{resnet} &89.63 {$\scriptstyle \pm 0.04 $}  &97.43 {$\scriptstyle \pm 0.19 $}   &72.29 {$\scriptstyle \pm  0.06$}  &78.74 {$\scriptstyle \pm  0.34$}  &66.78 {$\scriptstyle \pm  1.64$}  &84.97 {$\scriptstyle \pm  0.80$}  \\
        SENet154~\cite{hu2018squeeze}  &90.52{$\scriptstyle \pm  0.24$}  &97.85 {$\scriptstyle \pm 0.11 $}   &73.14 {$\scriptstyle \pm  0.58$}  &79.40 {$\scriptstyle \pm  0.51$}  & 67.44 {$\scriptstyle \pm  1.05$} &86.45 {$\scriptstyle \pm  0.88$} \\
        SwinV2-B~\cite{liu2022swin} &  90.82 {$\scriptstyle \pm  0.33$} & 98.21 {$\scriptstyle \pm  0.12$}& 72.78 {$\scriptstyle \pm  0.21$}& 79.84 {$\scriptstyle \pm  0.61$}& 67.67 {$\scriptstyle \pm  1.13$}& 86.42 {$\scriptstyle \pm  0.01$}\\
        \midrule
        PHG-Net~\cite{peng2024phg} &91.60 {$\scriptstyle \pm 0.43 $}  & 98.44 {$\scriptstyle \pm 0.23 $}  &74.57 {$\scriptstyle \pm  0.38$}  & 80.93{$\scriptstyle \pm 0.43 $} &69.00 {$\scriptstyle \pm 0.88 $}& 87.18 {$\scriptstyle \pm  0.26$} \\
        PHG-Net + multi-scale~\cite{gu2026integrating}  &92.32 {$\scriptstyle \pm 0.36$}  & 98.28 {$\scriptstyle \pm 0.27$}  &75.57 {$\scriptstyle \pm 0.70 $}  & \textbf{82.34} {$\scriptstyle \pm  0.72$} &70.00 {$\scriptstyle \pm 1.59 $}& 87.38 {$\scriptstyle \pm 0.02 $} \\
        \midrule
        Topo-SwinV2-B w/ PI  & {91.97} {$\scriptstyle \pm  0.31$}& {98.84} {$\scriptstyle \pm 0.08 $}& \textbf{75.62} {$\scriptstyle \pm 0.85 $}& {82.01} {$\scriptstyle \pm 0.38 $}& {69.92} {$\scriptstyle \pm  0.58$}& {87.23} {$\scriptstyle \pm 0.74 $}\\
       Topo-SwinV2-B w/ BC&{91.54} {$\scriptstyle \pm 0.50 $}  &{98.38} {$\scriptstyle \pm  0.20$}   &74.91 {$\scriptstyle \pm  0.47$}  & 81.54 {$\scriptstyle \pm  0.52$} &70.42 {$\scriptstyle \pm  0.62$}  &87.67 {$\scriptstyle \pm 0.29$} \\
       Topo-SwinV2-B w/ PS &\textbf{92.60} {$\scriptstyle \pm 0.27 $}  &\textbf{99.02} {$\scriptstyle \pm 0.06 $}   & 74.55 {$\scriptstyle \pm 0.22 $} &81.30 {$\scriptstyle \pm 0.63  $}  &\textbf{70.84} {$\scriptstyle \pm  0.49$}  &87.92 {$\scriptstyle \pm 0.43 $} \\
       Topo-SwinV2-B w/ PL &{91.35} {$\scriptstyle \pm  0.58$}  &{98.44} {$\scriptstyle \pm 0.13 $}   & 74.28 {$\scriptstyle \pm 0.24 $} & 80.89 {$\scriptstyle \pm 0.11 $} &70.05 {$\scriptstyle \pm  0.79$}  &\textbf{88.05} {$\scriptstyle \pm  0.57$} \\
        \bottomrule
    \end{tabular}
    }
\end{table}

\noindent
\textbf{Segmentation.}
Table~\ref{tab:seg} reports the segmentation results on the Synapse and ACDC datasets. We compare our Topo-Swin-UNet with two groups of methods: 2D segmentation methods including U-Net, TransUNet, Attention U-Net, and Swin-UNet, and 3D segmentation methods including V-Net, UNETR, Swin UNETR, nnFormer, and the recent UNETR++.
On Synapse, 2D methods generally under-perform compared to 3D methods since they
cannot exploit inter-slice volumetric context. Nevertheless,
Topo-Swin-UNet w/ PI achieves 82.19 Dice and 11.10 HD95, outperforming
all 2D baselines and remaining competitive with strong 3D methods. On
ACDC, Topo-Swin-UNet w/ PI achieves 91.73 Dice and the best HD95 of
0.84. Although UNETR++ and nnFormer obtain higher
Dice scores, our proposed method achieves
the lowest HD95 on ACDC, suggesting that the
topology-enhanced 2D model is especially effective for boundary
delineation.

\begin{figure}[t]
    \centering
    \includegraphics[width=0.94\linewidth]{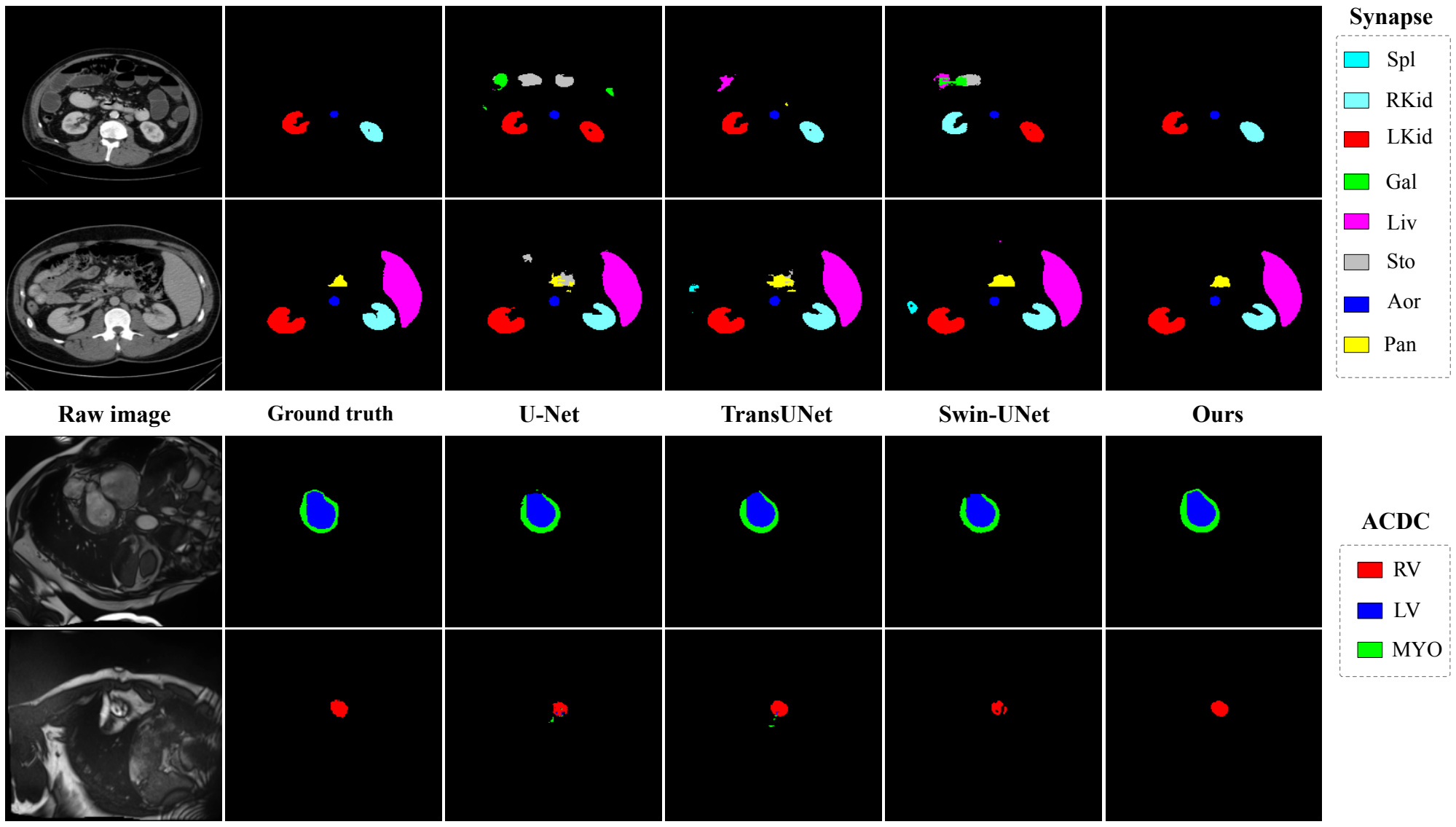}
    \caption{Visual segmentation results of different methods on the Synapse and ACDC datasets.}
    \label{vis:seg}
\end{figure}

Fig.~\ref{vis:seg} presents qualitative segmentation results on the Synapse and ACDC datasets. On Synapse, U-Net produces fragmented and anatomically inconsistent outcomes, particularly for small structures such as gallbladder (Gal) and stomach (Sto), which are either entirely missed or incorrectly localized. TransUNet improves boundary delineation but still exhibits spurious false-positive predictions in background regions. Swin-UNet produces cleaner segmentations but loses fine structural details for small organs. Our proposed method yields predictions most consistent with the ground truth across all organ classes, accurately recovering both large structures such as the liver (Liv) and spleen (Spl) and small challenging structures including the aorta (Aor) and pancreas (Pan), with sharper boundaries and fewer false positives. On the ACDC dataset, all competing methods show reasonable left ventricle (LV) segmentation due to its large size and high contrast; still, our proposed method demonstrates superior delineation of the myocardium (MYO) boundary and more precise right ventricle (RV) localization in the more challenging apical slice (the bottom row), where competing methods either miss the RV entirely or introduce spurious outputs. These qualitative results are consistent with the quantitative improvements reported in
Table~\ref{tab:seg}, and suggest that our proposed method learns more anatomically coherent and boundary-aware representations across both CT and MRI modalities.

\begin{table}[t]
    \centering
    \caption{Experimental results of different methods on the medical image segmentation datasets.}
    \label{tab:seg}
    \scalebox{1.0}{
    \begin{tabular}{lcccc}
        \toprule
        & \multicolumn{2}{c}{Synapse~\cite{landman2015miccai}} & \multicolumn{2}{c}{ACDC~\cite{bernard2018deep}} \\
        \cmidrule(lr){2-3}\cmidrule(lr){4-5}
         Method  & Dice $\uparrow$ & HD95 $\downarrow$ & Dice $\uparrow$ & HD95 $\downarrow$ \\
        \midrule
        U-Net~\cite{ronneberger2015u} &76.05 {$\scriptstyle \pm 0.39$}  &  19.02 {$\scriptstyle \pm 2.36$} &88.32 {$\scriptstyle \pm 0.70$}  & 3.58 {$\scriptstyle \pm 0.57$}  \\
        TransUNet~\cite{chen2021transunet} &77.34 {$\scriptstyle \pm 0.86$}  &16.83 {$\scriptstyle \pm 1.84$}   &89.42 {$\scriptstyle \pm 0.60$}  & 1.58 {$\scriptstyle \pm 0.14$}  \\
        Attention U-Net~\cite{oktay2018attention} &77.69 {$\scriptstyle \pm 0.44$}  & 17.29 {$\scriptstyle \pm 1.91$}  &86.82 {$\scriptstyle \pm 0.58$}  & 5.52 {$\scriptstyle \pm 0.42$}  \\
        Swin-UNet~\cite{cao2022swin} & 79.76 {$\scriptstyle \pm 1.23$} & 14.88 {$\scriptstyle \pm 1.60$} & 90.92 {$\scriptstyle \pm 0.77$} & 1.15 {$\scriptstyle \pm 0.65$} \\
        \midrule
        V-Net~\cite{milletari2016v}& 69.01 {$\scriptstyle \pm 0.92$} &33.69 {$\scriptstyle \pm 1.12$}   &84.74 {$\scriptstyle \pm 0.38$ } &5.15 {$\scriptstyle \pm 0.49$ }  \\
        UNETR~\cite{hatamizadeh2022unetr}& 78.98 {$\scriptstyle \pm 0.75$} &18.03 {$\scriptstyle \pm 1.55$}   &89.72 {$\scriptstyle \pm 0.36$} &2.33 {$\scriptstyle \pm 0.32$}  \\
        Swin UNETR~\cite{hatamizadeh2022swin}&82.62 {$\scriptstyle \pm 0.50$} &12.44 {$\scriptstyle \pm 1.06$}   & 91.60 {$\scriptstyle \pm 0.33$} &1.13 {$\scriptstyle \pm 0.18$}  \\ 
        nnFormer~\cite{zhou2021nnformer} &86.14 {$\scriptstyle \pm 0.48$} &11.98 {$\scriptstyle \pm 1.10$}  &92.34 {$\scriptstyle \pm 0.24$}  & 1.11 {$\scriptstyle \pm 0.35$} \\
        UNETR++~\cite{shaker2022unetr++} &\textbf{86.62} {$\scriptstyle \pm 0.42$} &\textbf{10.45} {$\scriptstyle \pm 1.44$}   & \textbf{92.98} {$\scriptstyle \pm 0.73$} &  0.98 {$\scriptstyle \pm 0.69 $} \\\midrule
        Topo-Swin-UNet w/ PI & {82.19} {$\scriptstyle \pm 0.29$}   & {11.10} {$\scriptstyle \pm 0.38$}& {91.73} {$\scriptstyle \pm 0.56$}& \textbf{0.84} {$\scriptstyle \pm 0.09$} \\
        Topo-Swin-UNet w/ BC & 79.84 {$\scriptstyle \pm 1.53$} &13.67  {$\scriptstyle \pm 0.96$} & 91.44 {$\scriptstyle \pm 0.54$}  & 1.00 {$\scriptstyle \pm 0.13$}  \\
        Topo-Swin-UNet w/ PS &81.38 {$\scriptstyle \pm 1.03$} &14.24 {$\scriptstyle \pm 1.39$}  & 91.65 {$\scriptstyle \pm 0.44$}& 0.86 {$\scriptstyle \pm 0.28$}  \\
        Topo-Swin-UNet w/ PL &80.28 {$\scriptstyle \pm 0.67$}  &14.64 {$\scriptstyle \pm 1.22$}  &91.47 {$\scriptstyle \pm 1.22$}  & 1.01 {$\scriptstyle \pm 0.18$}  \\
        \bottomrule
    \end{tabular}
    }
\end{table}

\noindent
\textbf{Effect of Topology Descriptor Choices.}
Descriptor performance is task-dependent for classification, with PS
performing the best on ISIC 2018, PI on CBIS-DDSM, and PS/PL on
RetinaMNIST. For segmentation, PI is the most consistent
descriptor across both datasets and achieves
the lowest HD95 among the evaluated
descriptors. Improvements across all four
descriptors suggest that the framework can
benefit from multiple persistent-homology
vectorizations rather than relying on a
single descriptor type. Detailed descriptor analysis is provided in
Appendix Sect.~\ref{appd:descriptor}.

\begin{table}[t]
    \centering
    \caption{Ablation study on CBIS-DDSM (classification) and Synapse
    (segmentation).}
    \label{tab:descriptor}
    \scalebox{0.92}{
    \begin{tabular}{lcccc}
        \toprule
        & \multicolumn{2}{c}{CBIS-DDSM} & \multicolumn{2}{c}{Synapse} \\
        \cmidrule(lr){2-3}\cmidrule(lr){4-5}
        Method   & Acc $\uparrow$ & AUC $\uparrow$ & Dice $\uparrow$ & HD95 $\downarrow$ \\
        \midrule
        SwinV2-B  / Swin-UNet    &72.78  &79.84  &79.76  & 14.88 \\
       Topo-SwinV2-B w/ PI / Topo-Swin-UNet w/ PI   &\textbf{75.62}  &{82.01}  &\textbf{82.19}  & \textbf{11.10} \\\midrule
       \rowcolor{sectiongray}
        \textit{Main Components Ablation} & & & & \\
        W/o Band-Aware H0/H1 Residual Refinement &73.58  &80.28  &81.03 &11.45  \\
        Fusion w/ concatenation  &72.73  & 78.99 &78.04 & 16.12 \\
        Fusion w/  attention-based residual gating   &73.15  & 80.22 &80.96 &12.95  \\\midrule
        \rowcolor{sectiongray}
        \textit{Band Selection Ablation} & & & & \\
        Short uniform-band across stages & 73.01 &79.92  &77.19 &17.82  \\
        Medium uniform-band across stages  & 74.86 &\textbf{82.19}  &81.14 &13.89  \\
        Long uniform-band across stages  & 74.43 &81.03  &81.34 &11.60  \\\midrule
       \rowcolor{sectiongray}
        \textit{Multi-scale Topology Representation Ablation} & & & & \\
        Single-scale at stage 1&74.01  &80.68  & 80.14 & 14.07 \\
        Single-scale and downsample &71.59  &78.42  &76.91 & 22.05 \\
        \bottomrule
    \end{tabular}
    }
\end{table}

\subsection{Ablation Studies}

Table~\ref{tab:descriptor} reports ablation studies on CBIS-DDSM and Synapse using PI as the topology descriptor.

\noindent
\textbf{Main Components Ablation.}
Removing the Band-Aware $H_0/H_1$ Residual Refinement module reduces
the accuracy by 2.04 points and Dice by 1.16 points, showing that raw
topology representations must be adapted to the downstream tasks.
Replacing the topology attention bias with concatenation causes the
largest drop: the accuracy decreases to 72.73, even below the non-topology
baseline (72.78), and Dice drops to 78.04, 4.15 points lower than the
full model.
This result suggests that explicit logit-level
topology modulation is more effective than the
evaluated feature-concatenation alternative
under the same experimental setting.
Replacing the topology bias with attention-based residual gating recovers
part of the performance (73.15 accuracy, 80.96 Dice), but still remains
well below the full model, highlighting the importance of injecting
topology directly into attention.

\noindent
\textbf{Band Selection Ablation.}
Compared with the evaluated uniform-band
strategies, progressive routing achieves
higher CBIS-DDSM accuracy and Synapse Dice,
as well as lower Synapse HD95, although
uniform medium-band routing attains a
slightly higher CBIS-DDSM AUC
(82.19 versus 82.01).
The uniform short-band strategy yields the
weakest results (73.01 accuracy, 77.19 Dice,
and 17.82 HD95). The uniform medium- and
long-band strategies perform better.
In particular, the uniform long-band strategy
achieves an HD95 of 11.60, approaching that
of the full model (11.10).
Nevertheless, progressive routing achieves
the strongest overall performance, supporting
its effectiveness as an architectural
inductive bias relative to the evaluated
uniform-band alternatives.



\noindent
\textbf{Multi-scale Topology Representation Ablation.}
Using only the Stage~1 topology representation reduces performance to
74.01 accuracy and 80.14 Dice, showing that fine-scale topology alone is
insufficient.
The largest drop occurs when the Stage~1 topology representation is
downsampled and reused for deeper stages instead of computing topology
natively at each scale: this reduces performance to 71.59 accuracy,
78.42 AUC, 76.91 Dice, and 22.05 HD95, falling even below the non-topology
baselines in all metrics.
These results demonstrate that the evaluated
fine-scale descriptor-downsampling strategy
is less effective than independently computing
persistent homology at each Swin stage's
spatial support. 
The full multi-scale topology representation construction achieves the best results in all
metrics and is the only variant that consistently improves over the
non-topology baselines across both tasks.

\subsection{Time and Parameter Complexity}

\textbf{Trainable Parameter Overhead.}
The new modules added to Topo-SwinV2-B introduce approximately
{83,300} trainable parameters on top of the 87.9M-parameter
Swin-V2-B backbone (i.e., an overhead of only {0.095\%}).
A full per-module breakdown is provided in Appendix Sect.~\ref{apped:param}.

\noindent
\textbf{Offline Multi-scale Topology Pre-computation Time.}
The multi-scale topology representations are computed once per dataset
split, stored on disk, and loaded as standard arrays during training;
no PD computation occurs at training time or during
evaluation on a fixed benchmark test set whose topology representations
have been pre-computed.
Using PS as the topology descriptor,
pre-computation requires {2h~16min} for RetinaMNIST (1,480 images,
${\approx} 5.5$s per image), and {7h~17min} in total for the
Synapse CT dataset (3,779 slices across training and test splits,
${\approx}7.0$s per image).
Note that pre-computation time can be reduced substantially by parallelizing
across images or scales. However, it does introduce a practical preprocessing latency at deployment. Thus, reducing the topology pre-computation cost is an important direction for
future research.

\section{Conclusions and Discussion}

In this paper, we proposed Topology-Reinforced Swin Transformer (TRiST), a new
topology-aware Swin-based model family for medical image analysis.
TRiST integrates topology into representation construction, refinement,
and token interaction with hierarchical multi-scale topology
representations, band-aware $H_0/H_1$ refinement, and persistence-band progressive attention bias injection. Across
five datasets, TRiST consistently improves over strong Swin-based
baselines while remaining competitive with topology-augmented methods.

The proposed method still has several limitations.
First, topology representations are pre-computed offline, which introduces
additional latency for deployment on unseen image data.
Second, the current framework is restricted to 2D $H_0/H_1$ features on
image patches and does not model native 3D topology.
Future work will focus on reducing the topology pre-computation cost and extending the
framework to native 3D PH for volumetric medical image analysis.

\section{Acknowledgements}
This research was supported in part by NSF grant CCF-2523787, AHA award 26AIREA1574568, and the Louisiana Board of Regents under Contract Number LEQSF(2025-28)-RD-A-21.
We also acknowledge the UTRGV High Performance Computing Resource.

\noindent\textbf{Competing Interests.}
The authors declare no competing interests.

\bibliographystyle{abbrvnat}
\bibliography{references}

\appendix
\newpage

\section{Datasets}
\label{apped:dataset}

We evaluate TRiST on three classification datasets and two segmentation
datasets.

\noindent
\textbf{(1) ISIC 2018~\cite{codella2019skin}.}
This skin lesion classification dataset contains 10,015 training images
and 193 validation images across 7 lesion classes. For fair comparison,
we follow the split setting in~\cite{zhuang2018skin} and report
5-fold cross-validation results.

\noindent
\textbf{(2) CBIS-DDSM~\cite{lee2017curated}.}
The Curated Breast Imaging Subset of the Digital Database for Screening
Mammography (CBIS-DDSM) contains 1,566 participants and 6,775 studies,
categorized as benign or malignant. We adopt the train/test split
provided by the dataset, using 20\% of the cases for testing and the
remaining cases for training.

\noindent
\textbf{(3) RetinaMNIST~\cite{medmnistv2}.}
This dataset contains 1,080 training images, 120 validation images, and
400 test images (1,600 images in total), each of size $224\times224$
pixels, categorized into five grades of diabetic retinopathy severity
(0--4). We follow the standard MedMNIST v2 train/validation/test split
and report classification accuracy on the official test set.

\noindent
\textbf{(4) Synapse CT~\cite{landman2015miccai}.}
This dataset contains 30 abdominal CT volumes with 8 annotated organs.
Following~\cite{chen2021transunet}, we split the dataset into 18 volumes
for training and 12 volumes for testing, corresponding to 2,211 training
slices and 1,568 test slices in our 2D setting. We report the average
Dice and 95\% Hausdorff distance (HD95) on eight organs: spleen (Spl),
right kidney (RKid), left kidney (LKid), gallbladder (Gal), liver (Liv),
stomach (Sto), aorta (Aor), and pancreas (Pan).

\noindent
\textbf{(5) ACDC~\cite{bernard2018deep}.}
This dataset contains 100 cardiac MRI cases for segmenting the left
ventricle (LV), right ventricle (RV), and myocardium (MYO). In total,
the dataset contains 1,902 2D image slices.  For fair comparison,
we report 5-fold cross-validation results.

\section{Implementation Details}
\label{appd:implementation}

All experiments are implemented in PyTorch. We use
Cubical Ripser~\cite{kaji2020cubical} to compute persistence diagrams.
Training is performed on a single NVIDIA A100 GPU (80\,GB) using the
AdamW optimizer~\cite{loshchilov2017decoupled} with weight decay
$0.005$. The input image size and patch size are set as $224\times 224$ and 4, respectively, following Swin Transformer V2~\cite{liu2022swin} and Swin-UNet~\cite{cao2022swin}.

For classification, the learning rate is $1\times10^{-4}$, the batch
size is 160, and the number of training epochs is 400. For
segmentation, the learning rate is $1\times10^{-4}$, the batch size is
24, and the number of training epochs is 200.

To preserve exact spatial alignment between the input image and the
pre-computed topology representations, we use only
topology-preserving data augmentation during training. Specifically, we
apply photometric transformations (e.g., color jitter and intensity
perturbation) and random horizontal/vertical flips, with flips applied
identically to both the image and its corresponding topology
representations. We do not use random resized cropping or arbitrary
rotations, since these would break the 1-to-1 alignment between
topology representations and the Swin token grid.

For ISIC 2018 and ACDC, we report mean $\pm$ standard deviation over
5-fold cross-validation. For CBIS-DDSM, RetinaMNIST, and Synapse, we
report mean $\pm$ standard deviation over 5 independent runs with
different random seeds.

\section{Stage-wise Region Sizes and Topology Representation Dimensions}
\label{appd:topo_scales}

\begin{table}[h]
\centering
\caption{Stage-aligned multi-scale topology representation construction
for the four-stage Swin hierarchy with $224\times224$ input images.}
\label{tab:topo_scales}
\begin{tabular}{cccccc}
\toprule
Stage $s$ & Token grid $H_s \times W_s$ & Stride $S_s$ & Region $R_s$ & Topology representation & Band \\
\midrule
1 & $56\times56$ & $4$  & $32\times32$   & $56\times56\times98$ & Short  \\
2 & $28\times28$ & $8$  & $64\times64$   & $28\times28\times98$ & Medium \\
3 & $14\times14$ & $16$ & $128\times128$ & $14\times14\times98$ & Long   \\
4 & $7\times7$   & $32$ & $192\times192$ & $7\times7\times98$   & Long   \\
\bottomrule
\end{tabular}
\end{table}

\section{Applications to Medical Image Classification and Segmentation}
\label{appd:application}

\paragraph{Topo-SwinV2-B for classification.}
Our classification model is built on Swin Transformer V2 Base~\cite{liu2022swin}
with embedding dimension $C=128$, stage depths $(2,2,18,2)$, attention
heads $(4,8,16,32)$, and window size $8\times8$. At each stage, the
pre-computed topology representation $\mathbf{P}_s$ is normalized,
refined, decomposed, and converted into a stage-specific topology bias,
which is injected into every block of that stage.

In addition to topology-guided attention, we use a lightweight topology
bypass branch to preserve global topology cues. Specifically, the
refined topology representations from Stages~2--4 are spatially
average-pooled, projected by three independent branch MLPs,
concatenated, and fused into a global topology summary vector
$\mathbf{f}_{\mathrm{topo}}\in\mathbb{R}^{128}$. Let
$\mathbf{f}_{\mathrm{vis}}\in\mathbb{R}^{1024}$ denote the global visual
feature after the backbone and global average pooling. The final
prediction is computed from the concatenated visual and topology
summaries:
\[
    \hat{\mathbf{y}}
    =
    \mathbf{W}_{\mathrm{head}}
    \bigl[
        \mathbf{f}_{\mathrm{vis}};
        \mathbf{f}_{\mathrm{topo}}
    \bigr]
    +
    \mathbf{b}_{\mathrm{head}}.
\]
All newly introduced topology parameters are
trained from scratch, while the SwinV2-B
backbone is initialized from ImageNet-pretrained
weights. The zero-initialized topology scoring
output ensures that attention-logit injection
has no effect at initialization, while the
zero-initialized residual MLP outputs limit
the initial refinement updates.


\paragraph{Topo-Swin-UNet for segmentation.}
For segmentation, we use the same topology construction, refinement, and
topology-bias modules within a Swin-UNet~\cite{cao2022swin}
encoder--decoder architecture. The encoder follows the standard
multi-stage Swin hierarchy, and the topology bias is injected into the
window-attention blocks at the corresponding resolutions. In the
decoder, topology biases are reused by resolution so that encoder and
decoder blocks operating at the same spatial scale share the same
topology prior. This design keeps the parameter overhead low while
maintaining stage-consistent topology guidance across the full
segmentation network.

\section{Trainable Parameter Breakdown}
\label{apped:param}

Table~\ref{tab:param} provides a per-module breakdown of the trainable
parameters added to Topo-SwinV2-B relative to the 87.9M-parameter
Swin-V2-B backbone~\cite{liu2022swin}. The
\texttt{PersistenceBandDecomposer} contains no learnable parameters,
since it performs only fixed index slicing over the 98-dimensional
descriptor. The classification-head expansion adds exactly $128K$ weight
parameters, where $K$ is the number of output classes, amounting to 896
(ISIC 2018, $K=7$), 256 (CBIS-DDSM, $K=2$), and 640
(RetinaMNIST, $K=5$). The bias vector of size $K$ is already present in
the baseline classification head and therefore does not constitute an
additional parameter overhead.

\begin{table}[h]
    \centering
    \caption{Trainable parameter breakdown for the topology modules added
    to Topo-SwinV2-B. The Swin-V2-B backbone has 87,930,848 parameters.
    The total topology overhead is 0.095\% of the backbone.}
    \label{tab:param}
    \scalebox{0.695}{
    \begin{tabular}{lrr}
        \toprule
        Module & Parameters & Details \\
        \midrule
        LayerNorm $\times 4$ (98-dim)
            & 784
            & $2 \times 98 \times 4$ \\
        BandAwareH0H1ResidualRefinement $\times 4$
            & 26,328
            & 6 MLPs (hidden dim.\ 32) + 6 gates + 2 row gates per stage \\
        PersistenceBandDecomposer
            & 0
            & fixed index slicing, no learnable parameters \\
        PersistenceBandWindowAttentionBias $\times 4$
            & 12,224
            & scoring MLP
              $(\mathrm{Linear}(d_{\mathrm{band}},64)\rightarrow\mathrm{GELU}\rightarrow\mathrm{Linear}(64,h_s))$
              + $\beta_s$ per stage \\
        TopologyBypassSummary
            & 43,968
            & $3\times\mathrm{Linear}(98,64)$ + $\mathrm{Linear}(192,128)$ + $\mathrm{LayerNorm}(128)$ \\
        Head expansion ($+128$ input dim)
            & $\leq 896$
            & $128\times K$, with $K \leq 7$ \\
        \midrule
        \textbf{Total topology overhead}
            & \textbf{$\approx 83{,}300$}
            & \textbf{0.095\% of backbone} \\
        \bottomrule
    \end{tabular}
    }
\end{table}

\section{Time and Parameter Complexity}
\label{appd:complexity}

\paragraph{Trainable parameter overhead.}
The topology modules added to Topo-SwinV2-B introduce approximately
\textbf{83,300} trainable parameters on top of the 87.9M-parameter
Swin-V2-B backbone, corresponding to an overhead of only
\textbf{0.095\%}. This includes four
\texttt{BandAwareH0H1ResidualRefinement} instances (26,328 parameters),
one \texttt{TopologyBypassSummary} (43,968 parameters), four
\texttt{PersistenceBandWindowAttentionBias} modules (12,224 parameters),
four \texttt{LayerNorm} layers (784 parameters), and a small head
expansion ($\leq 896$ parameters). A full per-module breakdown is given
in Table~\ref{tab:param}.

\paragraph{Offline topology pre-computation time.}
The multi-scale topology representations are computed once per dataset
split, stored on disk, and loaded as standard arrays during training.
Thus, no persistence diagram computation occurs during training or
during evaluation on a fixed benchmark test set whose topology
representations have already been pre-computed. However, a new unseen
test image still requires persistent homology computation before
inference. Therefore, TRiST does not add trainable complexity to the
backbone, but it does introduce a practical preprocessing latency at
deployment.

Using persistence statistics (PS) as topology descriptor,
pre-computation requires \textbf{2h 16min} for RetinaMNIST (1,480 images,
approximately 5.5\,s per image) and \textbf{7h 17min} in total for the
Synapse CT dataset (3,779 slices across training and test splits,
approximately 7.0\,s per image). This cost can be reduced substantially
by parallelizing across images or scales, and reducing this bottleneck
is an important direction for future work.

\begin{figure}[!t]
    \centering
    \includegraphics[width=1.0\linewidth]{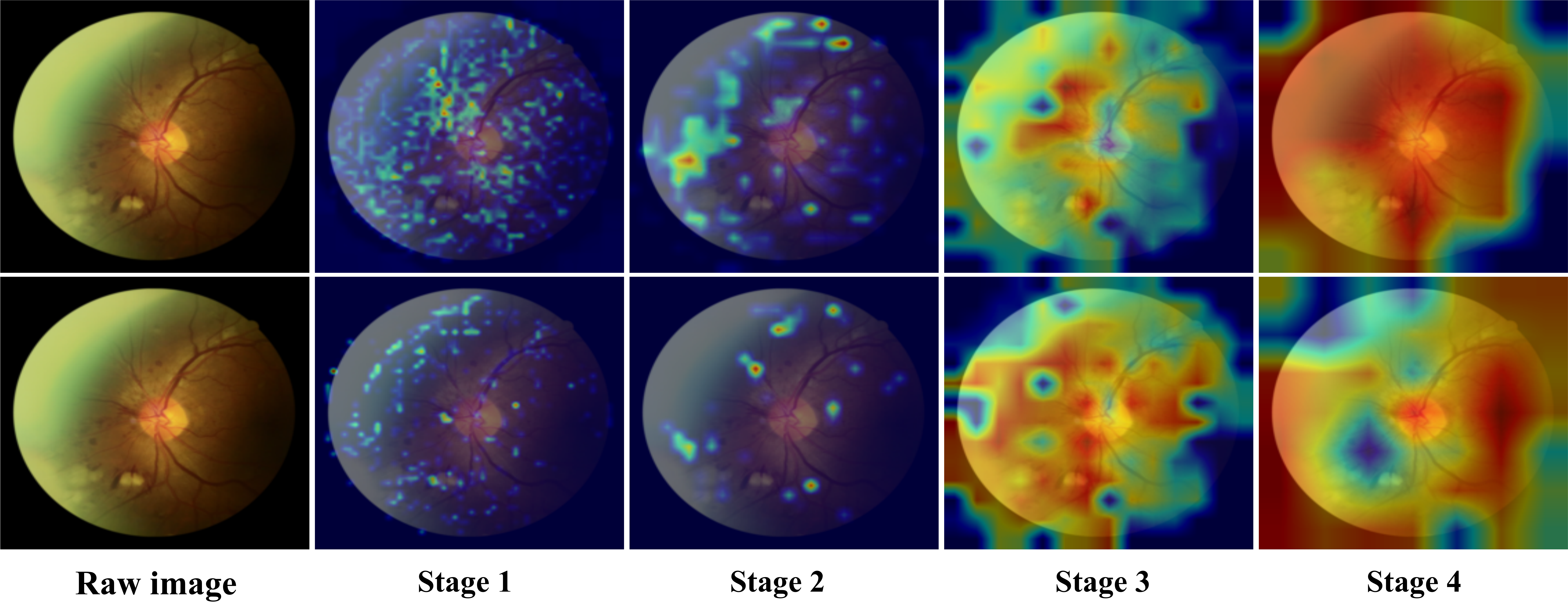}
    \caption{GradCAM visualizations comparing Topo-SwinV2-B w/ PS (top
    row) and SwinV2-B (bottom row) on a Severe DR (Grade~3) fundus
    image. Columns show the raw image followed by activation maps from
    Stage~1 to Stage~4. Warmer colors indicate stronger GradCAM
attribution for the target prediction.}
    \label{vis:class}
\end{figure}

\section{Qualitative Results}
\label{apped:qual}

\paragraph{Classification.}
Fig.~\ref{vis:class} presents GradCAM
visualizations across four stages for
SwinV2-B and Topo-SwinV2-B w/ PS on
a Grade~3 diabetic retinopathy fundus
image. The two models exhibit different
spatial attribution patterns across the
hierarchical feature representations.

The observed differences suggest that
topology reinforcement influences the
learned representations. However,
GradCAM does not directly establish
correspondence between persistence
generators, learned topology scores,
and clinically annotated anatomical
structures. These results should
therefore be interpreted as qualitative
evidence of representational differences
rather than direct validation of
anatomical localization or clinical
reasoning.


\section{Effect of Topology Descriptor Choices}
\label{appd:descriptor}

On classification, descriptor performance is task-dependent. PS achieves
the best accuracy and AUC on ISIC 2018, suggesting that compact global
persistence statistics provide sufficient discriminative signal while
also acting as a natural regularizer. On CBIS-DDSM, PI provides the
strongest overall performance among the TRiST variants, achieving the
best accuracy, while PHG-Net + multi-scale attains the highest AUC
overall. This suggests that PI better preserves the structural cues
needed for TRiST on mammography classification. On RetinaMNIST, PS gives
the best accuracy, while PL gives the best AUC, indicating that the
ranked functional encoding of PL is well suited to subtle differences
across adjacent diabetic retinopathy severity grades.

For segmentation, PI performs most consistently across both datasets. On
Synapse, PI achieves the best Dice (82.19) and HD95 (11.10) among all
Topo-Swin-UNet variants, outperforming BC, PS, and PL by a clear
margin. This is consistent with the nature of dense prediction, where PI
retains both the spatial location and lifetime of topological events,
providing richer structural information for boundary localization. On
ACDC, all four descriptors obtain similar Dice scores (91.44--91.73),
but PI again achieves the lowest HD95 (0.84), indicating better boundary agreement among the evaluated descriptors. 

Importantly, all four descriptors outperform
the corresponding non-topology baselines,
suggesting that the proposed framework can
benefit from multiple persistent-homology
vectorizations rather than relying on a
single descriptor type. The contributions
of hierarchical alignment, refinement,
and topology-guided attention are examined
separately through the component ablations.

\newpage

\section*{NeurIPS Paper Checklist}

\begin{enumerate}

\item {\bf Claims}
    \item[] Question: Do the main claims made in the abstract and introduction accurately reflect the paper's contributions and scope?
    \item[] Answer: \answerYes{}{} 
    \item[] Justification: The abstract and introduction clearly state the main claims and contributions of the paper, including hierarchical multi-scale topology representation, Band-Aware $H_0/H_1$ Residual Refinement, and Persistence-Band Progressive Attention Bias. These claims are supported by experiments on five medical imaging benchmarks.

\item {\bf Limitations}
    \item[] Question: Does the paper discuss the limitations of the work performed by the authors?
    \item[] Answer: \answerYes{} 
    \item[] Justification: The paper discusses key limitations, including the offline pre-computation cost of topology descriptors and the current restriction to 2D $H_0/H_1$ topology rather than native 3D persistent homology.

\item {\bf Theory assumptions and proofs}
    \item[] Question: For each theoretical result, does the paper provide the full set of assumptions and a complete (and correct) proof?
    \item[] Answer: \answerNA{} 
    \item[] Justification: This paper does not make formal theoretical claims requiring theorems or proofs. The contribution is methodological and empirical.

    \item {\bf Experimental result reproducibility}
    \item[] Question: Does the paper fully disclose all the information needed to reproduce the main experimental results of the paper to the extent that it affects the main claims and/or conclusions of the paper (regardless of whether the code and data are provided or not)?
    \item[] Answer: \answerYes{} 
    \item[] Justification: The paper describes the proposed architecture, multi-scale topology representation computation algorithm, refinement module, attention-bias construction, datasets, and evaluation protocol in sufficient detail for reproduction. Additional implementation details are provided in the Appendix.

\item {\bf Open access to data and code}
    \item[] Question: Does the paper provide open access to the data and code, with sufficient instructions to faithfully reproduce the main experimental results, as described in supplemental material?
    \item[] Answer: \answerNo{} 
    \item[] Justification: We do not release code at submission time. However, all datasets used in this work are publicly available from their original sources, and the paper and appendix provide detailed descriptions of the model, training protocol, and evaluation settings.

\item {\bf Experimental setting/details}
    \item[] Question: Does the paper specify all the training and test details (e.g., data splits, hyperparameters, how they were chosen, type of optimizer) necessary to understand the results?
    \item[] Answer: \answerYes{} 
    \item[] Justification: The paper specifies the datasets, model variants, evaluation metrics, and experimental protocol. Additional details, including data splits and implementation settings are provided in the Appendix.

\item {\bf Experiment statistical significance}
    \item[] Question: Does the paper report error bars suitably and correctly defined or other appropriate information about the statistical significance of the experiments?
    \item[] Answer: \answerYes{} 
    \item[] Justification: Results are reported with measures of variability (mean $\pm$ standard deviation) over repeated 5 runs or 5 cross-validation folds, depending on the dataset.

\item {\bf Experiments compute resources}
    \item[] Question: For each experiment, does the paper provide sufficient information on the computer resources (type of compute workers, memory, time of execution) needed to reproduce the experiments?
    \item[] Answer: \answerYes{} 
    \item[] Justification: The appendix reports compute-related details, including GPU hardware and software package for computing 2D topology.
    
\item {\bf Code of ethics}
    \item[] Question: Does the research conducted in the paper conform, in every respect, with the NeurIPS Code of Ethics \url{https://neurips.cc/public/EthicsGuidelines}?
    \item[] Answer: \answerYes{} 
    \item[] Justification: We have read the NeurIPS Code of Ethics and believe that this work complies with it.

\item {\bf Broader impacts}
    \item[] Question: Does the paper discuss both potential positive societal impacts and negative societal impacts of the work performed?
    \item[] Answer: \answerNo{} 
    \item[] Justification: The current submission does not include a dedicated discussion of broader societal impacts beyond methodological limitations and intended medical-imaging research use.
    
\item {\bf Safeguards}
    \item[] Question: Does the paper describe safeguards that have been put in place for responsible release of data or models that have a high risk for misuse (e.g., pre-trained language models, image generators, or scraped datasets)?
    \item[] Answer: \answerNA{} 
    \item[] Justification: This work does not release a high-risk generative model or other system requiring special safeguards.

\item {\bf Licenses for existing assets}
    \item[] Question: Are the creators or original owners of assets (e.g., code, data, models), used in the paper, properly credited and are the license and terms of use explicitly mentioned and properly respected?
    \item[] Answer: \answerNo{} 
    \item[] Justification: The paper cites the original sources of the datasets and baseline methods used in the experiments, but it does not explicitly document license or reuse terms for these assets.

\item {\bf New assets}
    \item[] Question: Are new assets introduced in the paper well documented and is the documentation provided alongside the assets?
    \item[] Answer: \answerNA{} 
    \item[] Justification: This submission does not release a new public dataset, benchmark, or other public asset.

\item {\bf Crowdsourcing and research with human subjects}
    \item[] Question: For crowdsourcing experiments and research with human subjects, does the paper include the full text of instructions given to participants and screenshots, if applicable, as well as details about compensation (if any)? 
    \item[] Answer: \answerNA{} 
    \item[] Justification: This work does not involve crowdsourcing or human-subject experiments conducted by the authors.

\item {\bf Institutional review board (IRB) approvals or equivalent for research with human subjects}
    \item[] Question: Does the paper describe potential risks incurred by study participants, whether such risks were disclosed to the subjects, and whether Institutional Review Board (IRB) approvals (or an equivalent approval/review based on the requirements of your country or institution) were obtained?
    \item[] Answer: \answerNA{} 
    \item[] Justification: This work does not involve human-subject experimentation requiring IRB approval.

\item {\bf Declaration of LLM usage}
    \item[] Question: Does the paper describe the usage of LLMs if it is an important, original, or non-standard component of the core methods in this research? Note that if the LLM is used only for writing, editing, or formatting purposes and does \emph{not} impact the core methodology, scientific rigor, or originality of the research, declaration is not required.
    \item[] Answer: \answerNA{} 
    \item[] Justification: Large language models were not used as an important, original, or non-standard component of the core methodology, experiments, or scientific contribution of this research.

\end{enumerate}

\end{document}